\documentclass[conference]{IEEEtran}
\IEEEoverridecommandlockouts
\usepackage{cite}
\usepackage{amsmath,amssymb,amsfonts}
\usepackage{algorithmic}
\usepackage{graphicx}
\usepackage{textcomp}
\usepackage{xcolor}
\def\BibTeX{{\rm B\kern-.05em{\sc i\kern-.025em b}\kern-.08em
    T\kern-.1667em\lower.7ex\hbox{E}\kern-.125emX}}

\usepackage{graphics} 
\usepackage{epsfig} 
\usepackage{mathptmx} 
\usepackage{times} 
\usepackage{amsmath} 
\usepackage{amssymb}  
\usepackage{caption}
\usepackage{subcaption}
\usepackage{hyperref}

\title{\LARGE \bf
Precise Object Placement Using Force-Torque Feedback
}

\author{\IEEEauthorblockN{Osher Lerner, Zachary Tam, Michael Equi}
\IEEEauthorblockA{\textit{Department of Electrical Engineering and Computer Science} \\
\textit{University of California at Berkeley}\\
Berkeley, CA 94720, USA \\
{\tt\small \{oshlern, ztam, michaelequi\}@berkeley.edu}}}

\begin{document}

\maketitle
\thispagestyle{empty}
\pagestyle{empty}

\begin{abstract}

Precise object manipulation and placement is a common problem for household robots, surgery robots, and robots working on in-situ construction. Prior work using computer vision, depth sensors, and reinforcement learning lacks the ability to reactively recover from planning errors, execution errors, or sensor noise. This work introduces a method that uses force-torque sensing to robustly place objects in stable poses, even in adversarial environments. On 46 trials, our method finds success rates of 100\% for basic stacking, and 17\% for cases requiring adjustment.

\end{abstract}

\section{INTRODUCTION}



Stable object stacking is a common task for home robots, industrial/warehouse robots, and exploring robots. For example, robots may be required to stack dishes in a sink, cabinet, or dishwasher, or stably stack tools and parts on a shelf. Furthermore, when exploring cities after a natural disaster, or when exploring other planets, the ability to construct impromptu shelter from provided rubble is critical.

When humans stack objects, the last mile of placement is guided primarily by fine-grain force sensing and reactive control. This is especially apparent when sight is removed or when trying to precariously balance complex objects. In such actions, humans empirically tend to lightly release the grasped object in order to sense the small net object wrench and compensate accordingly.

In this work, we use an ATI Axia80 wrist-mounted force-torque sensor to emulate this behavior. We are able to sense when the robot’s net wrench from the grasped object is zero, which by Newton’s Third Law means that the robot is applying zero net wrench to the grasped object. At this point, the object can safely be released.
This known stable placement can be useful for rock stacking, robust placement and stacking of delicate objects, and human-robot handover. The work considers extensions of the grasp matrix/friction cone paradigm of grasping presented in lectures in the latter half of the course.

Previous work in robotic stone stacking \cite{c4, c5, c6} uses vision based sensing to find stable stacking placements, however it generally lacks the ability to recover from visual input noise and robot kinematic error. This work seeks to apply the force-torque feedback methods commonly used for peg insertion \cite{c1, c2} to stone stacking in order to enable contact-based readjustment for stable stacking.

In applying force-torque sensing to stone stacking, we attempt to minimize accidental motion of the grasped object and the existing structure, while simultaneously exploring the stacking pose space to find one with the optimal wrench. We thus present a policy that estimates a gradient for pose optimization to reduce external torques to zero, i.e., to find a stable stacking pose.

\section{RELATED WORK}


\subsection{Force-Torque Feedback}
Wrist-mounted force torque sensors such as the ATI Axia80 are very common for high precision and contact-critical tasks such as peg insertion and deburring \cite{c1}, \cite{c2}, \cite{c3}. \cite{c1} uses impedance control and system force/torque modeling to determine and zero out peg reaction forces. \cite{c2} extends prior work to a deformable pegboard; because of the nonlinear dynamics, the authors use deep reinforcement learning to model the problem. \cite{c3} uses fuzzy PI control in order to maintain strong contact with the desired part for deburring. For peg insertion tasks, the authors of \cite{c1} and \cite{c2} consider only the last mile of contact adjustments, starting from a vision-based initial guess. In this work, we take a similar approach of starting from a good initial contact and making force-torque based adjustments in order to reach our goal. However, we apply these concepts to the problem of stacking, where the goal is less well-defined and it is important to minimize disturbances to the environment.
\subsection{Stone Stacking}
Prior work on stone stacking uses various types of vision-based sensing to find optimal stacks. \cite{c4} uses an objects with known 3D models and then simulates possible stacks to find a stable one to execute. \cite{c5} uses a Q-learning approach for stone stacking where object models are known and the problem is reduced to 2D by using simple extruded polygons as stones. It tests with physical run-throughs of simulated rollouts. \cite{c6} uses deep reinforcement learning to stack 3D objects of varying irregularity in simulation from a ground-truth depth map of the bottom of the held object and the top of the existing structure. \cite{c7} uses vision sensing and sim-to-real transfer of a reinforcement learning agent to form especially challenging stacks of known polyhedra and cylinders. This paper extends these works by applying force-torque sensing to accommodate for noise and inaccuracies in vision sensing and object modeling.
\subsection{Alternative Stacking}
We also note prior use of a novel end effector for grasping and stacking objects. \cite{c8} includes a demonstration of a novel ``bubble'' gripper which senses finger normal and shear forces in order to detect grasped object geometry and external forces. This work demonstrates similar flat object stacking using a simpler and more robust parallel-jaw gripper.

\section{METHODS}



  \begin{figure}[thpb]
      \centering
      \includegraphics[width=8cm]{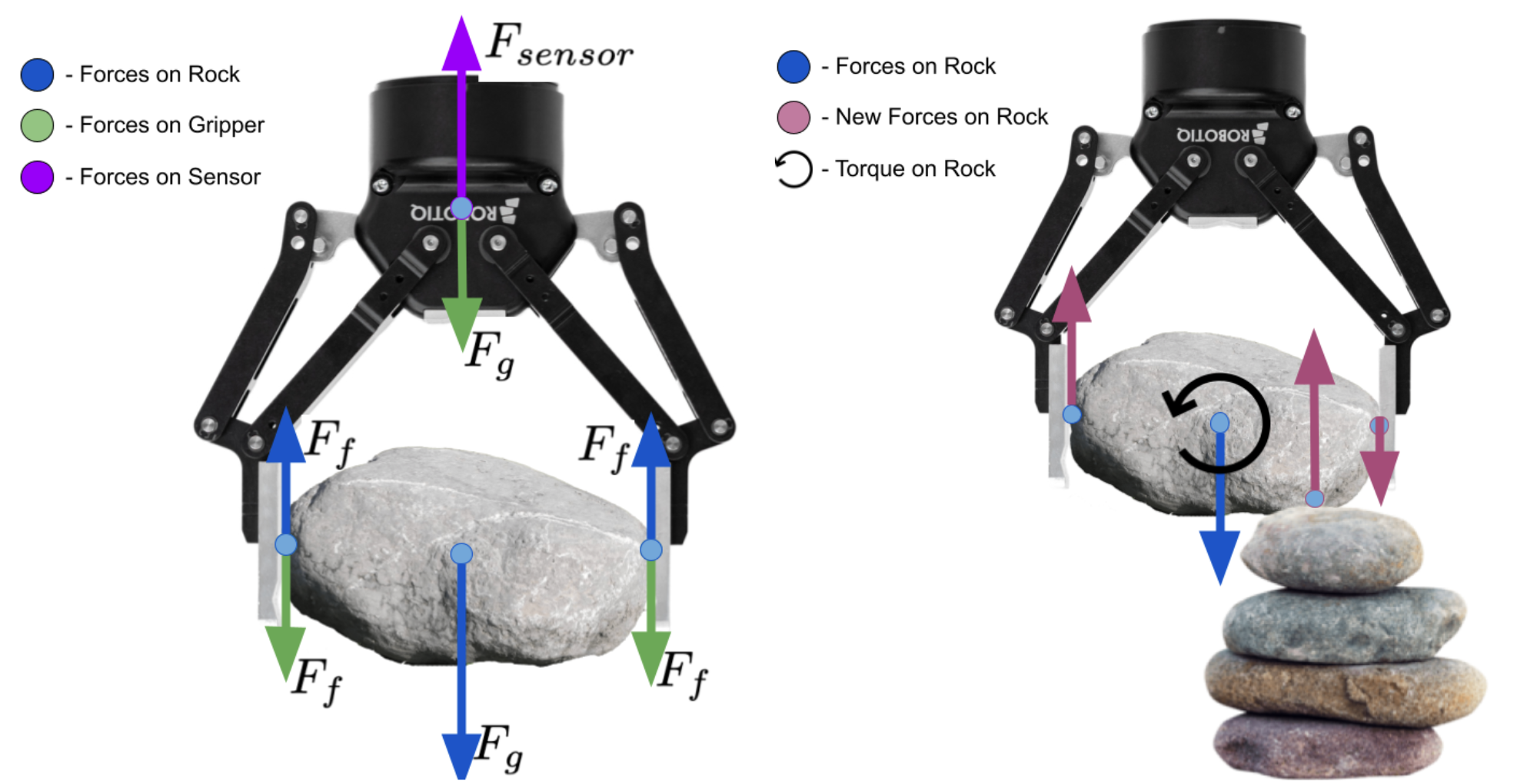}
      \caption{Left: The force diagram of a grasp before placement (normal forces at the jaws omitted for clarity). Right: The new forces on the rock during a place attempt, and the resulting torque.}
      \label{grip_diagram}
  \end{figure}

  
\subsection{Pipeline}

We devise an algorithm to pick, place, and iteratively adjust placement until the rock can be released stably. We first implement a standard pick-and-place algorithm to pick a rock from the ground and form an initial guess for placement. The end effector is quasi-statically lowered onto this pose until we detect significant resistance from the force-torque sensor.
After halting in this configuration, we use the force-torque readings to calculate the approximate point of contact and its surface normal. We then slightly raise the rock to relieve the forces and adjust the pose to place the center of mass of the rock above the previous point of contact, and optionally further navigate along the surface towards a flatter region. This process repeats until the torques detected at the time of contact are sufficiently small, and we can release the rock in a stable configuration.

This algorithm is orchestrated through ROS. The robotic arm receives position and velocity commands in the world frame, and converts them into a joint-space control loop, respecting physical and safety constraints on position, speed, and acceleration. We use a node for computer vision to estimate the position of rocks on the ground and an initial guess for placement on the tower, but currently spoof the outputs with an approximate oracle. These outputs are varied adversarially to measure the performance of the rest of the pipeline. The force torque sensor node constantly publishes readings, and internally handles calibration and reference frame conversions. Finally the gripper node takes calibration, open, and close commands.


\subsection{Physics Model}\label{physics}

For the workspace frame, we choose the robot's base frame, labeled $B$. We denote the wrist frame by $W$, the gripper-tip frame by $T$, the rock's COM (center of mass) frame by $R$, and the contact point frame as $C$.

The rock's true COM is unknown, so we approximate it to be directly at the center of the parallel jaws: $g_{RT} = I$. Given a better estimate of the COM, we'd simply have to adjust our measurements by an additional frame conversion. (One possible extension of this project would be to use force-torque readings to estimate the COM of the object.)

\begin{figure}[thpb]
      \centering
      \includegraphics[width=9cm]{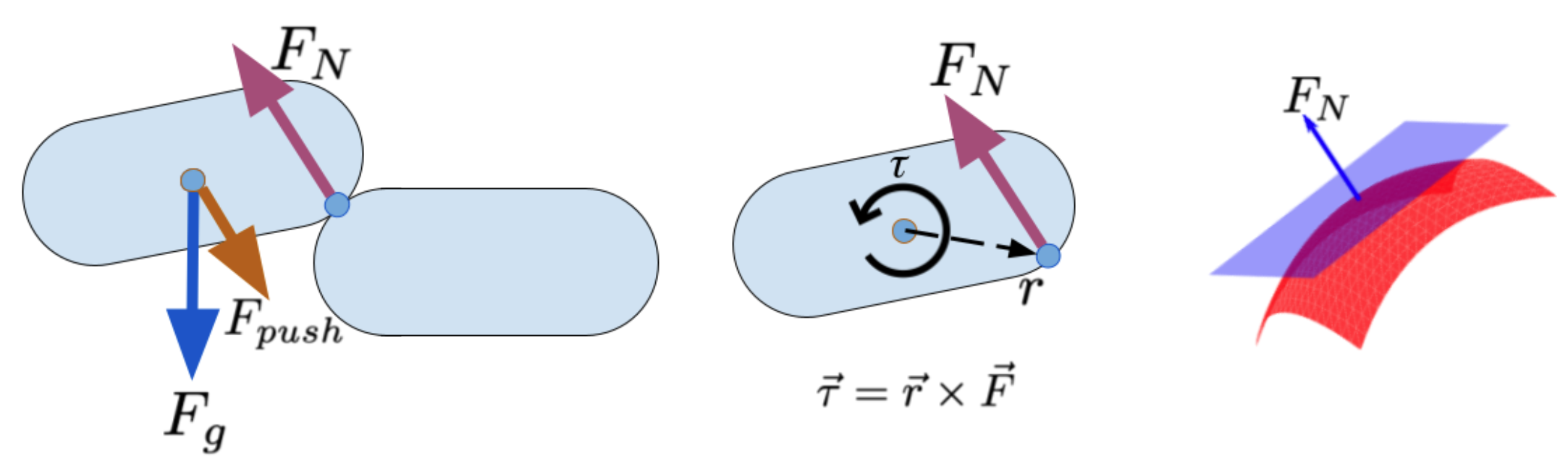}
      \caption{Left: The simplified force diagram of an attempted rock placement, used to compute the normal force at rest. Middle: The torque induced on the rock COM, which we reverse engineer to find the contact point r from a measured torque. Right: The normal force is along the surface normal of the tower. Since the tower is likely convex around the point of contact, this can be used to find the direction towards a flat peak.}
      \label{contact_diagram}
\end{figure}

As we lower the rock, it makes contact with the tower. We label the point of contact (relative to the rock's COM) by $\vec{r}$. At the point of contact, the tower exerts a normal force $\vec{F}_N$ on the rock in the direction of the surface normal (Fig. \ref{contact_diagram}). This results in a torque on the rock's COM given by $\vec{\tau} = \vec{r} \times \vec{F}_N$. 

In general, a wrench in frame $A$ can be specified as a vector concatenation of torque and force
$\mathcal{F}_A = \begin{bmatrix} \vec{\tau} \\ \vec{F} \end{bmatrix}$. Assuming a rigid body connects $A$ to a frame $B$ (related by $g_{AB} \in SE(3)$), the wrench felt in frame $B$ can be computed as \cite{c9}

\begin{equation}\label{eq:adj}
    \mathcal{F}_B = \text{Ad}_{g_{AB}}^T \mathcal{F}_A
\end{equation}

As long as our rigidity assumptions hold (e.g. no slipping), the wrench on the rock is sensed as a force and torque on the wrist by our sensor. We convert this reading to the rock COM frame using (Eq. \ref{eq:adj}) and a pre-measured transformation between the sensor and the gripper tip frames.

We wish to reconstruct the position of the contact point and orientation of the surface from our sensor readings. Even though we derived how to compute the wrench felt by the rock, this wrench is a product of 2 unknowns: the contact point $\vec{r}$ and the normal force $\vec{F}_N$. To solve this, we leverage our quasi-static assumption to claim the net force on the rock is $\approx 0$ when we measure.

The forces acting on the rock are gravity, the push of the gripper, and the normal force from the tower (see Fig. \ref{contact_diagram}). The gravitational force we can measure while holding the rock in a hovering position. The push of the gripper we computed by converting the wrench from the force torque sensor. This means we can derive the normal force from 
\begin{equation}
    \vec{F}_{net} = \vec{F}_g + \vec{F}_{push} + \vec{F}_N = 0.
\end{equation}

By computing $\vec{F}_N$, we know the direction of the surface normal of the tower at the contact point. Now we seek $\vec{r}$, knowing $\vec{\tau} = \vec{r} \times \vec{F}_N$. We can recover 
\begin{equation}
    \vec{r}_T = \frac{1}{|\vec{F}_N|^2} \vec{F}_N \times \tau
\end{equation}
where $\vec{r}_T$ is the component of $\vec{r}$ within the tangent plane of the tower surface. The component normal to the surface cannot be deduced it from measuring forces and torques. Fortunately, for the purpose of selecting a new placement we need only shift our pose along the tower surface tangent plane, since any other pose would either not make contact with the tower or be inside it.

\subsection{Surface Normal Optimization}

The above approach for computing the displacement to the point of contact is used in our optimization algorithm to select the next placement such that the rock center of mass is directly over the previous contact point. For complex objects, finding a balanced pose requires not only centering the COM above the contact point, but locating a good contact point at which the contact surface is flat.

To this end, we utilize our calculation of the surface normal. However, finding a flat contact point on the tower fundamentally involves higher order information about the surface: curvature describes the rate of change of the surface normal. Since we gain no curvature information from our measurement process, we must look to some useful assumptions. The point of contact is almost certainly extremity, since it is the first point on the surface of the tower we reach on a linear approach, particularly since we take a vertical approach and the objects we handle are wide rocks. Around extremities, the surface is convex, meaning it curves away in every direction. Since we seek the direction along the surface where the normal vector is closest to vertically upwards, the best guess is therefore the direction within the tangent plane that has the largest z-component. Mathematically, using the normal vector $\hat{n} = \vec{F}_N/|\vec{F}_N|$ this direction is given as 
\begin{equation}
    \hat{d} = - \hat{n} \times (\hat{n} \times \hat{z})
\end{equation}
We can then simply add a small step in the direction of $\hat{d}$ to $\vec{r}_T$ to determine our proposed shift towards a more stable configuration.

\subsection{Calibration}\label{calib}

Calibration is thoroughly critical to the success of this approach. We go about calibration in two phases. The first measures the forces and torques on the wrist when the object is grasped and suspended statically by the robot. The second measures the forces and torques on the wrist when putting the object onto a flat surface. Taking a baseline measurement of the forces and torques when pushing the object onto a flat surface, such as the location where the object was picked up, gives us a reference point for when the object is in a stable position. We can then subtract this reference from the forces and torques when stacking to determine when the object is close enough to the stable reference to be released without falling. The primary purpose of the first measurement is to find the weight of the object / gripper combination which is used to separate out the normal force as a result of gravity and the force with which the robot is pushing down on the object.

\begin{figure}
\centering
    \begin{subfigure}{4cm}
      \centering
      \includegraphics[width=4cm]{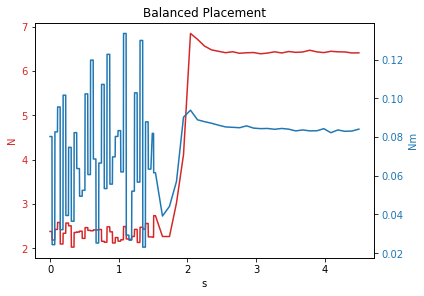}
    \end{subfigure}%
    \begin{subfigure}{4cm}
      \centering
      \includegraphics[width=4cm]{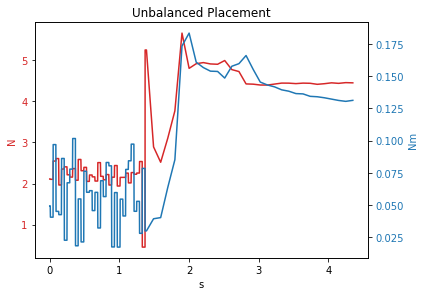}
    \end{subfigure}
\caption{Norms of force and torque readings upon contact during vertical descent. Left: Centered placement. Right: Off-center placement.}
\label{FT_data}
\end{figure}

When taking each force-torque measurement we wait for 0.1 seconds after the motion of the arm has stopped and take an additional 0.5 seconds of samples which are then averaged to devise our final force-torque reading at that point in time. Without this step, our readings would be overwhelmed by noise and the 25Hz read-rate of the sensor. This is to help improve the consistency of the measurements and thus result in faster convergence to stable placement positions. (See Fig. \ref{FT_data})

Rigidity is central to our mathematical derivations, yet slippage is commonplace between the object and the gripper as well as the object and the tower. We designed our algorithm for consistency since our physical assumptions were often violated. By approaching placement vertically each time, we avoid slippage along the surface of contact. This also serves to zero out deviations from our model due to misalignment of normal forces, gravity, and the arm's pushing forces. Due to imprecise grasping of the objects and the nonlinear fashion in which our gripper pose depends on the object it is gripping, our dynamical models failed to accurately compute terms and convert between frames, but these complications are zeroed out when everything aligns with the direction of gravity.

\section{RESULTS}\label{results}

\subsection{Physical Setup}
In this work, we use a 2014 UR5 CB3 Cobot equipped with an ATI Axia80 force-torque sensor and a Robotiq 2F-140 parallel-jaw gripper. Since this paper is concerned primarily with the placement section of pick and place, we define a preset picking process. As a stand-in for a vision-based object detection and grasping algorithm, a human operator places objects in a specified pickup location on flat ground. The gripper is able to repeatedly collect objects from this loading zone and place them in a specified initial stacking position. The initial stacking position is a hard-coded stand-in for an optimization warm-start we could otherwise source from an algorithm like the one presented in \cite{c6}. The existing stack structure is aligned with this initial position before experiments begin.

Due to the difficulties we faced in the later parts of this work, we stack with largely flat objects such as shuffleboard pucks (which have flat bottoms but only undulating semi-flat tops) and flat squares of 1/4 inch fiberboard.

In these experiments, our inputs were the initial position of the existing stack and the selected object to place, while our output was the success of an attempted stack placement (a binary variable). We hypothesized that more complex objects and existing stacks would result in lower likelihood of a successful stack and more update steps before a stable pose was found.

\subsection{Stacking Experiments}
\begin{figure}[thbp]
      \centering
      \includegraphics[width=8cm]{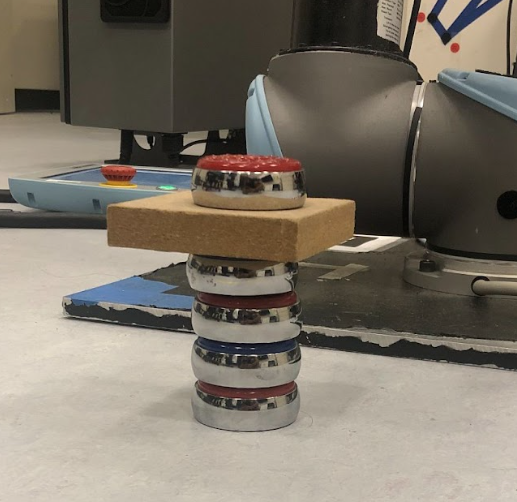}
      \caption{A six-object stack constructed from the ground-up by our policy.}
      \label{stack_easy}
 \end{figure}

Using the setup described above, we performed 46 documented stacking tests. On the 16 trials where no update steps are required (i.e., the initial stacking pose corresponds to a stable placement), our policy detects this with 100\% accuracy and precision. This is successful even though the algorithm has no information on how high the existing stack is (other than an upper bound). We were able to stack at least 6 objects before we ran out of objects to stack in the lab space (Fig. \ref{stack_easy}).

However, we find only a 17\% success rate on more difficult experiments.

On trials where the placement surface is non-horizontal, the algorithm will not succeed unless we allow the robot to adjust the orientation of the end-effector (not accounted for in this work). Instead, it makes the inaccurate implicit assumption that the ramp leads to a stable placement at the top, thus never finding a stable stacking pose on the ramp.

Even with near-horizontal stable stacking poses, when multiple steps are required for stacking we also find limited success (18\%). In practice, we find the gradient update steps to be inconsistent at moving the object toward a stable pose, and we find that the metric for finding a stable placement pose is often misaligned with the real world.

\subsection{Algorithm Component Experiments and Analysis}
Due to the challenges we faced in actual stacking experiments, we performed other experiments and analysis with smaller components of our algorithm. These results serve to demonstrate the individual success and theoretical feasibility of these components, and help to narrow down the source of any remaining error.

 
In early development before the ROS architecture properly communicated, we conducted experiments to manually analyze and visualize the force-torque readings from place actions. This procedure proved insightful later and throughout the project to diagnose the algorithm's performance. We list example data from three attempted placements of a held puck unto a puck on the ground. The readings are calibrated with the object suspended in the air, and processed to eliminate remaining offsets.

\begin{figure}[thpb]
      \centering
      \includegraphics[width=8cm]{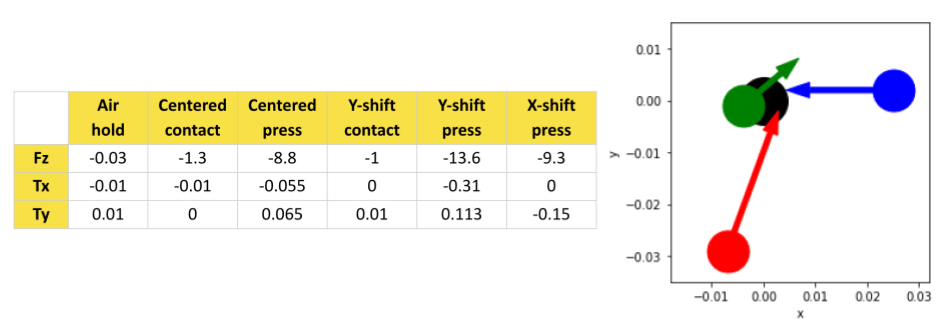}
      \caption{Left: Experimental readings of characteristic states from 3 place maneuvers. Upon first contact, noise still overwhelms torque readings. After pressing with 10N, the direction of the stable site can be inferred from the data. 
      Right: Plotted center of mass for each candidate pose and the algorithm's suggested shift. "Centered" in green, "Y-shift" in red, "X-shift" in blue, and true tower center in black.}
      \label{place_data}
 \end{figure}

In the first experiment, we spoof a "guessed" placement pose from vision to be quite close to the true center of the puck below it. The second is offset mostly in the y direction, and the third is offset mostly in the x direction. The data from characteristic states during placement is listed in (Fig \ref{place_data}), and the initial placements relative to the tower center (0,0) are plotted. Once sufficiently pressed against the tower, we plug the force torque readings into our algorithm to propose candidate shifts. The proposed shifts are remarkably successful, pointing to locations within 1cm of the optimal placement in just one iteration.

However, we notice the proposed shift for the already nearly centered placement is an arbitrary direction, leading to a worse guess for the next attempt. We observe this failure mode consistently throughout our experiments: update steps are accurately computed for egregiously offset placements, but as we near a stable configuration the algorithm fails to converge and proposes further poses in random directions.

When the pipeline failed to place objects, we would create manual contact forces to verify the accuracy of our readings, calibration, and update steps. This process is done by simply applying a strong pressure to the object with the experimenter's finger while the robotic arm lowers it towards the tower. We were able to get clear readings with up to 30Nm of torque, and the proposed shift directions passed this sanity check 100\% of the time.



\subsection{Implications}
Given the independent success of our placement algorithm and our update steps, both in simulation and in sanity checking, we are able to narrow down the source of the error when we combine the subcomponents of our method.

While the finger pressing sanity check demonstrates the success of our algorithm for high forces, it says little about how well it scales down (i.e., for the very small torques when the current pose is only marginally off from a stable pose). Furthermore, stacking tests where update steps were not required seem to show that the algorithm is able to detect stable poses. Therefore, we suspect a sensitivity limit to our algorithm, i.e., it cannot detect the direction of small torques with enough accuracy to get the object into a stable pose while simultaneously being able to distinguish a stable pose from one that is unstable but nearby. This explains the observed pattern of behavior of accurate adjustments from extremely unstable placements, then incorrect adjustments from partially stable poses.

\section{CONCLUSION AND FUTURE WORK}
\addtolength{\textheight}{-3cm}   
                                  
In this work, we were able to use solely force-torque sensing to consistently stack reasonably flat objects in a given location and to occasionally find stable stacks from an unstable initial pose.

When working on this project, we encountered two major difficulties. Firstly, we were trying to implement novel software on three new independent pieces of hardware. Ultimately, setup took a considerable amount of time, as did tuning the force-torque sensor (Section \ref{calib}). Once we had set up the hardware, we encountered significant challenges with getting proper force-torque readings in order to find accurate gradient steps (see Section \ref{results}).

The first steps for future work are to incorporate better methods for detecting small offsets from stable poses; this is where we found difficulty in this work. In order to leverage our successful approach into consistent convergence, we must modify our calculations for nearly-balanced poses. 

One possibility is to slowly release the objects in order to predict how they might behave if fully released. We can thus determine whether to release or where to move the object next.

Once we are able to consistently stack flat rocks, we plan to stack real rocks which are bulky, irregular, and rough. This comes with nonconvexity, multiple-contact points, and ill-posed optimization. For these cases we can utilize our surface normal optimization edition to our procedure. Although our derivation assumed many good conditions, we believe it will succeed on most common objects, particularly due to our resistance threshold and the nature of placing. 

For objects that slip into unstable configurations, or generally rotated pick and place actions, we need to account for more than just vertical motion. This requires consideration of the gripper model center of mass and the wrench it applies to the force-torque sensor. It also requires optimizing over a higher dimensional space, where we must consider out-of-plane torques and changes in gripper orientation as optimization steps.

For multi-rock sequential stacking, we can record the final position of the center of mass of each placed rock and use it as our initial guess for the next rock's placement. When our control loop fails to converge, we place the rock on the ground and grasp the last rock on the tower using its stored pose. We remove it as well, and try a random new sequence of rock stacks, until those are exhausted and we backtrack further down the tower. This search doubles as a reordering of rocks as well as just random repositioning due to the noise in our stacking algorithm. 

A major challenge we encountered was with the stickiness of the gripper fingers. In order to press against the environment without losing grip of the target object, high friction is required for the gripper fingers. However, this comes with stickiness, which tends to cause the grasped object to move upon release (i.e., it gets stuck to the gripper fingers). This problem is especially apparent with the slick shuffleboard pucks.

A promising future approach to tackling the complex nonlinear nature of this optimization procedure is to leverage deep learning. Some assumptions have to be made to make progress in the pick-and-place task, and deep learning enables us to implicitly learn the relevant assumptions from patterns in data. This can be beneficial for both general agents with hard-to-specify assumptions and specialized datasets for particular objects and environments. Today's cutting edge research \cite{c5} uses reinforcement learning, which our method could utilize to form strategies for sequentially sampled placements; incorporating an RL agent could enable interaction with the environment with the intent of information gathering to quicken long-term success, as opposed to only attempting successful placement with the present action. A deep model could also potentially hybridize the tasks of vision and tactile sensing, leveraging the relationship between the two more than a separated pipeline could.

Alternatively, a 0th order sampling of higher order information is possible: by sampling readings at a local array of points, we can interpolate gradients towards flat and balanced positions.


\section{ACKNOWLEDGMENTS}

The authors gratefully acknowledge the contribution of National Research Organization and reviewers' comments. We would also like to thank Professor Ruzena Bajcsy, Isabella Huang, and Ellis Ratner for generously allowing us to borrow their equipment and laboratory space.


\end{document}